\documentclass[conference]{IEEEtran}
\IEEEoverridecommandlockouts

\usepackage{cite}
\usepackage{amsmath,amssymb,amsfonts}
\usepackage[noend]{algorithmic}
\usepackage{graphicx}
\usepackage{textcomp}
\usepackage{xcolor}
\usepackage{booktabs}
\usepackage{multirow}
\usepackage{microtype}
\usepackage{hyperref}
\usepackage{cleveref}
\usepackage{todonotes}
\usepackage{algorithm}
\usepackage{subcaption}

\def\BibTeX{{\rm B\kern-.05em{\sc i\kern-.025em b}\kern-.08em
    T\kern-.1667em\lower.7ex\hbox{E}\kern-.125emX}}
\begin{document}

\title{Personalized Dynamic Difficulty Adjustment -- \newline Imitation Learning Meets Reinforcement Learning}

\IEEEoverridecommandlockouts
\IEEEpubid{\makebox[\columnwidth]{ 979-8-3503-5067-8/24/\$31.00~\copyright2024 IEEE \hfill} 
\hspace{\columnsep}\makebox[\columnwidth]{ }}

\author{\IEEEauthorblockN{Ronja Fuchs, Robin Gieseke, and Alexander Dockhorn}
\IEEEauthorblockA{\textit{Faculty of EECS} \\
\textit{Leibniz University Hannover}\\
Hannover, Germany \\
dockhorn@tnt.uni-hannover.de} 
}

\maketitle

\begin{abstract}
Balancing game difficulty in video games is a key task to create interesting gaming experiences for players. 
Mismatching the game difficulty and a player's skill or commitment results in frustration or boredom on the player's side, and hence reduces time spent playing the game. 
In this work, we explore balancing game difficulty using machine learning-based agents to challenge players based on their current behavior.
This is achieved by a combination of two agents, in which one learns to imitate the player, while the second is trained to beat the first.
In our demo, we investigate the proposed framework for personalized dynamic difficulty adjustment of AI agents in the context of the fighting game AI competition.
\end{abstract}


\section{Introduction}

Engaging players with an appropriate challenge is fundamental to a rewarding gaming experience. A study by Hagelback and Johansson~\cite{hagelback2009measuring} has shown that static difficulty settings often fail to adapt to individual skill levels, leading to frustration for beginners or boredom for veterans. At the same time, players reported games against an opponent that adapts to their performance to be more enjoyable. Dynamic Difficulty Adjustment (DDA) techniques~\cite{zohaib2018dynamic} address this issue by adjusting difficulty based on a player's performance, therefore, aiming to keep them in a constant state of flow~\cite{csikszentmihalyi1990flow}.

This paper proposes a novel DDA technique, which aims to train a personalized opponent based on the player's current behavior. While deep reinforcement learning (DRL) has been shown to result in human-compatible performance in a variety of gaming tasks~\cite{shao2019survey}, it is known to converge slowly, previously making it unsuitable for a real-time scenario.
To overcome this issue, we implement a two-step process, in which one agent is asked to imitate the player's behavior, and a second agent is trained to compete against the first. At specific intervals, the player's current opponent is replaced with the second agent, presenting the player with a customized challenge. The replacement occurs after a specified number of observation steps. For this initial study, we swapped the agents after each single step in order to be suitable for short play sessions.

In the following, we shortly present related work on DDA (\Cref{sec:background}) before introducing our proposed agent model~(\Cref{sec:method}). A preliminary evaluation~(\Cref{sec:experiments}) has shown promising results in terms of reported player satisfaction. \Cref{sec:conclusion} shortly concludes our work.
Our related demo aims to present the proposed approach to the conference participants, allowing us to openly discuss the model while collecting more data on user experiences.

\section{Background and Related Work on DDA}
\label{sec:background}

To achieve a state of flow~\cite{csikszentmihalyi1990flow}, the developers' objective is to provide players with sufficiently challenging activities that result in interesting and meaningful interactions~\cite{hunicke2005case}. 
DDA approaches often trade the designers' control of the experience with the game's freedom in creating new and interesting challenges on the fly.  

In this work, we expand on previous approaches, which heavily focus on generating personalized levels~(e.g. \cite{yannakakis2011experience}). We not only parameterize an existing AI agent but propose a method to learn an AI opponent's behavior from scratch. 
Designers are freed from creating flexible AI opponents but must still identify player performance variables and game mechanics affecting difficulty, as outlined by Hendrix et al.~\cite{hendrix2018implementing}.

\section{Personalized Dynamic Difficulty Adjustment}
\label{sec:method}

To allow for a personalized dynamic difficulty adjustment (PDDA) experience that adapts to the player's skill, we propose a framework combining imitation learning and reinforcement learning agents.
Seamless integration of our PDDA machine learning-based opponent is achieved through a combination of three agents:
\begin{itemize}
    \item \textbf{Opponent Agent}: The agent the player is currently playing against, which is to be replaced seamlessly while playing the game.
    \item \textbf{Imitation Learning Agent}: An agent that observes the player's behavior and learns to replicate its actions.
    \item \textbf{Reinforcement Learning Agent}: An agent trained to win against the imitation learning agent.
\end{itemize}

\begin{figure}[t]
    \centering
    \includegraphics[width=1.0\columnwidth]{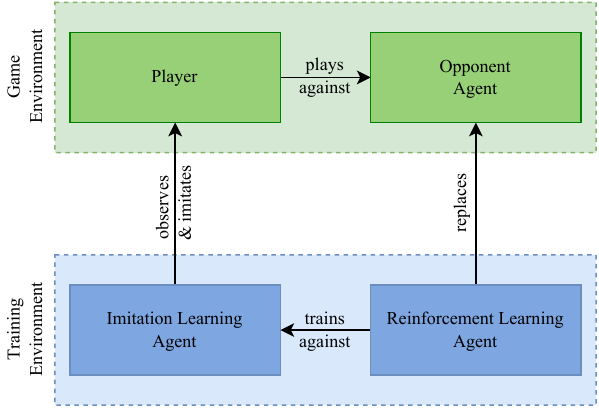}
    \caption{Proposed agent architecture for personalized dynamic difficulty adjustment.}
    \label{fig:agent-model}
\end{figure}

Starting with a simple rule-based \textbf{opponent agent}, we collect observations on the player's actions. The opponent agent is replaced with the current reinforcement learning agent at fixed intervals. Longer intervals will result in less frequent changes in the opponent, giving the training process more time. This can result in stronger agents to compete against but comes at the cost of less dynamic interactions.

Once enough observations have been stored, the \textbf{imitation learning agent} is trained to reproduce the player's actions. For the development of our prototype, we used the Python-based library \textit{River}\footnote{https://pypi.org/project/river/} due to its implementation of an adaptive random forest classifier~\cite{gomes2017adaptive} that ensures fast learning every time the data set is updated. At the same time, it automatically detects drifts in the training data distribution, allowing the model to adapt to changes in player behavior. 

For the implementation of the \textbf{reinforcement learning agent}, our prototype makes use of the Advantage Actor Critic algorithm~\cite{mnih2016asynchronous} and its synchronous and deterministic implementation by Stable Baselines 3~\cite{stable-baselines3}. We train the agent in a separate training environment running in the background. 
 
\section{Preliminary Evaluation of PDDA}
\label{sec:experiments}

This study serves as an initial exploration into the feasibility of our approach. To evaluate it, we implemented the proposed PDDA model in the context of the FightingICE framework~\cite{lu2013fighting}. 

Throughout our study, the imitation learning agent received the relative position of both characters and the actions they currently perform, as well as the player's input.
When predicting player actions, we achieved an accuracy of  82-87~\% on the training set with the imitation learning agent (we report a range of values due to the stream mining setup). 

For the reinforcement learning agent's input, we chose to use the current screen encoded as a $96\times 64$ grayscale image. Here, we augmented the image by encoding the characters' hit points and energy levels in the top left and top right corners of the screen using the grayscale value of two pixels each. For the sake of simplicity, we used the implementation's default convolutional neural network (CNN) policy~\footnote{\url{https://stable-baselines3.readthedocs.io/en/master/_modules/stable_baselines3/common/policies.html}}.
During training, we gave the agent a positive reward for decreasing the other character’s hit points and a penalty for losing its own.

It's important to note that due to time constraints, our study was conducted with a sample size of 5 participants, who each played three games against the MCTS agent provided by the framework and the proposed agent model. 
After every game, they had to rate their experience on a scale from 1-10. Players on average reported higher ratings while playing against the proposed agent model (7.0$\pm 1.09$) when compared to the MCTS agent (6.6$\pm 1.01$). 
We are aware that the system's expressiveness is limited and requires further testing, i.e. to evaluate the impact of each model's accuracy on the players' experience.

\section{Conclusion and Future Work}
\label{sec:conclusion}
The framework for personalized dynamic difficulty adjustment presented in this work provides the opportunity to challenge players according to their individual skill levels. The combination of an imitation learning agent and a reinforcement learning agent results in a personalized system for DDA that requires minimal setup by the designers.
In the future, we aim to expand on the evaluation of the proposed PDDA by increasing the number of study participants and analyzing the system's impact on their perceived difficulty and player satisfaction for different opponent agents. Furthermore, the implementation of more complex imitation learning agents could provide a better approximation of the player's behavior and may result in stronger opponents to match more experienced players.
Extending this work to high-difficulty games, as well as exploring the accuracy of the agent to meaningfully generalize to the human player is a future consideration.


\bibliographystyle{unsrt}
\bibliography{references.bib}

\end{document}